%% file: main.tex
\documentclass{sigchi}

\toappear{\scriptsize Permission to make digital or hard copies of all or part of this work for personal or classroom use is granted without fee provided that copies are not made or distributed for profit or commercial advantage and that copies bear this notice and the full citation on the first page. Copyrights for components of this work owned by others than the author(s) must be honored. Abstracting with credit is permitted. To copy otherwise, or republish, to post on servers or to redistribute to lists, requires prior specific permission and/or a fee. Request permissions from permissions@acm.org. \\
{\emph{CHI '20, April 25--30, 2020, Honolulu, HI, USA.} } \\
Copyright is held by the owner/author(s). Publication rights licensed to ACM. \\
ACM ISBN 978-1-4503-6708-0/20/04\ ...\$15.00.\\
http://dx.doi.org/10.1145/3313831.3376523}

\input{packages}

\newcommand{\tool}{RoomShift}

\newcommand{\final}{1}

\newcommand {\changes}[1]{{\color{purple}{#1}\normalfont}}
\newcommand {\ryo}[1]{{\color{purple}\bf{Ryo: #1}\normalfont}}
\newcommand {\daniel}[1]{{\color{green}\bf{Daniel: #1}\normalfont}}
\newcommand {\warning}[1]{{\color{purple}\bf{[#1]}\normalfont}}
\newcommand {\nothing}[1]{}
\newcommand {\hooman}[1]{{\color{blue}\bf{[#1]}\normalfont}}

\ifthenelse{\equal{\final}{1}}
{
\renewcommand {\changes}[1]{{#1}}
\renewcommand {\ryo}[1]{}
\renewcommand {\daniel}[1]{}
\renewcommand {\warning}[1]{}
\renewcommand {\hooman}[1]{}
}
{}

\begin{document}
\title{
\tool{}: Room-scale Dynamic Haptics for VR\\ with Furniture-moving Swarm Robots
\vspace{-0.2cm}
}

\numberofauthors{1}
\author{
  \alignauthor{
    Ryo Suzuki~$^1$,
    Hooman Hedayati~$^1$,
    Clement Zheng~$^2$, 
    James Bohn~$^1$,\\
    Daniel Szafir~$^{1,2}$,
    Ellen Yi-Luen Do~$^{1,2}$,
    Mark D. Gross~$^{1,2}$, 
    Daniel Leithinger~$^{1,2}$\\
  \affaddr{
    University of Colorado Boulder,\\
    $^1$Department of Computer Science,
    $^2$ATLAS Institute
  }\\
  \email{
    \{ryo.suzuki, hooman.hedayati, clement.zheng, james.bohn\\
    daniel.szafir, ellen.do, mdgross, daniel.leithinger\}@colorado.edu
  }}\\
}

\teaser{
\vspace{-0.4cm}
\centering 
\includegraphics[height=2.86cm]{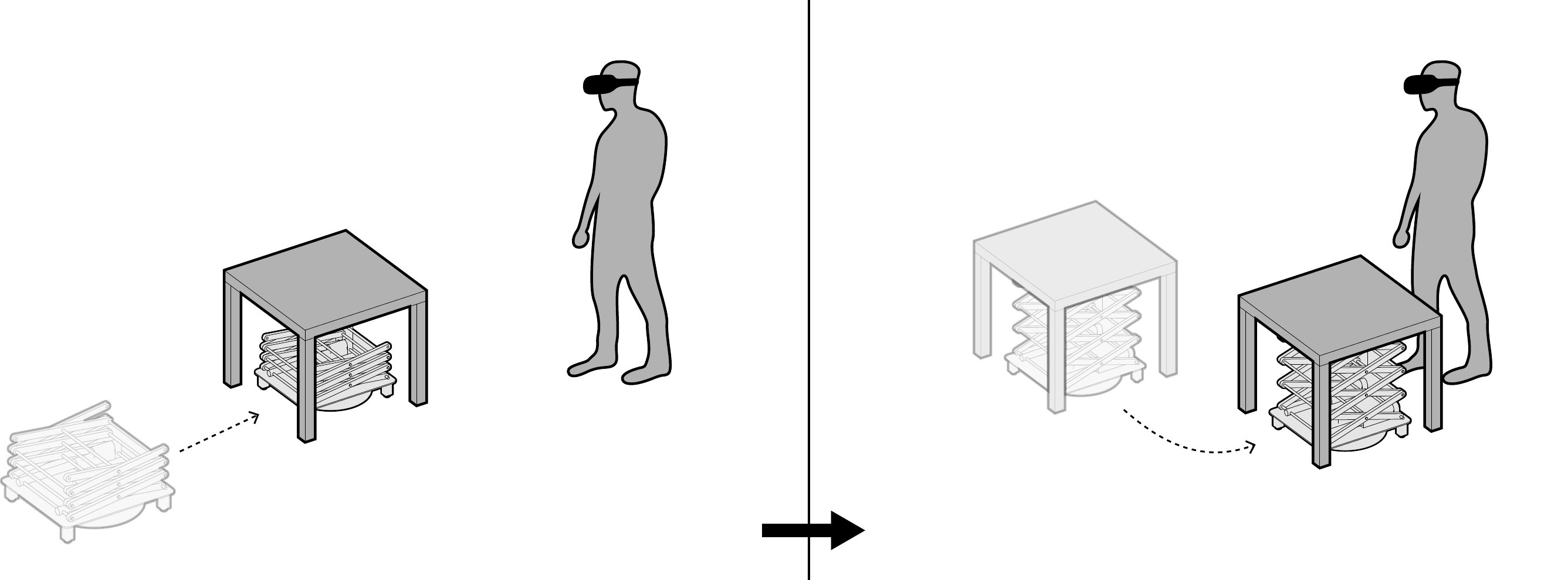}
\includegraphics[height=2.86cm]{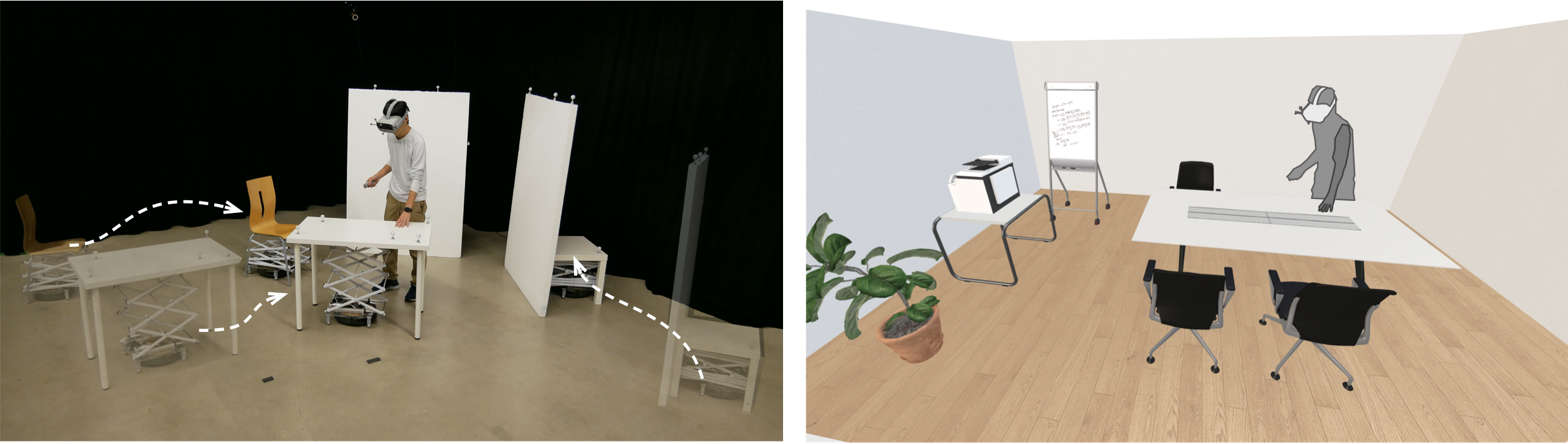}
\caption{
\tool{} is composed of a swarm of shape-changing robots for haptic feedback in VR. RoomShift robots move beneath a piece of furniture to lift, move and place it. Multiple robots move furniture to construct a physical haptic environment collectively. The corresponding virtual scene is shown, with a human silhouette added for a reference.
}
\label{fig:cover}
\vspace{-0.3cm}
}

\maketitle

\input{0-abstract.tex}
\input{1-introduction.tex}

\input{2-related-work.tex}
\input{4-system.tex}

\input{5-applications.tex}

\input{7-limitations.tex}

\input{8-conclusion.tex}
\input{acknowledgements.tex}

\balance
\bibliographystyle{SIGCHI-Reference-Format}
\bibliography{references}

\end{document}

%% file: packages.tex
\usepackage{balance}
\usepackage{graphics}
\usepackage[T1]{fontenc}
\usepackage{txfonts}
\usepackage{mathptmx}
\usepackage[pdflang={en-US},pdftex]{hyperref}
\usepackage{color}
\usepackage{booktabs}
\usepackage{textcomp}
\usepackage{microtype}
\usepackage{ccicons}
\usepackage{todonotes}
\usepackage{listings}
\usepackage{cleveref}

\def\plaintitle{RoomShift}

\def\emptyauthor{}
\def\plainkeywords{}

\makeatletter
\def\url@leostyle{%
  \@ifundefined{selectfont}{
    \def\UrlFont{\sf}
  }{
    \def\UrlFont{\small\bf\ttfamily}
  }}
\makeatother
\def\pprw{8.5in}
\def\pprh{11in}

\definecolor{linkColor}{RGB}{6,125,233}
\hypersetup{%
  pdftitle={\plaintitle},
  pdfauthor={\emptyauthor},
  pdfkeywords={\plainkeywords},
  pdfdisplaydoctitle=true, 
  bookmarksnumbered,
  pdfstartview={FitH},
  colorlinks,
  citecolor=black,
  filecolor=black,
  linkcolor=black,
  urlcolor=linkColor,
  breaklinks=true,
  hypertexnames=false
}

\usepackage{enumitem}

%% file: 0-abstract.tex
\begin{abstract}
RoomShift is a room-scale dynamic haptic environment for virtual reality, using a small swarm of robots that can move furniture. RoomShift consists of nine shape-changing robots: Roombas with mechanical scissor lifts. These robots drive beneath a piece of furniture to lift, move and place it. By augmenting virtual scenes with physical objects, users can sit on, lean against, place and otherwise interact with furniture with their whole body; just as in the real world. When the virtual scene changes or users navigate within it, the swarm of robots dynamically reconfigures the physical environment to match the virtual content. We describe the hardware and software implementation, applications in virtual tours and architectural design and interaction techniques. 
\end{abstract}

\ccsdesc[500]{Human-centered computing~Human computer interaction (HCI)}
\printccsdesc
\keywords{haptic interfaces; room-scale haptics; virtual reality; swarm robots}
\vspace{0.5cm}



%% file: 1-introduction.tex
\section{Introduction}
There is a clear need to provide haptic sensations in virtual environments.
Recent advances in display and tracking technologies promise immersive experience in virtual reality, but objects seen in VR such as walls and furniture are only visual: the user cannot touch, feel, sit on, or place objects on them. This limits the sense of full immersion in the virtual world. 
To overcome these limitations, various haptic interfaces have been explored.
In the previous work, most haptic interfaces focus on finger-tip haptic feedback \changes{with actuated controllers}~\cite{benko2016normaltouch, choi2017grabity, choi2016wolverine} or on-body haptic sensations \changes{with wearable devices}~\cite{lindeman2006wearable, nagai2015wearable, delazio2018force, lopes2015impacto, schmidt2015level}.
\changes{In contrast, encountered-type} haptic feedback with a dynamic environment promises to increase the immersion of virtual experiences~\cite{cheng2014haptic, cheng2015turkdeck,teng2019tilepop,vonach2017vrrobot}, \changes{which are difficult to achieve using an only handheld or wearable haptic devices}.
Through a dynamic haptic environment, \changes{users can touch and interact with the whole virtual scene with their bodies} --- they can walk, sit on, and lean against objects in the VR environment.
Existing approaches \changes{for actuated environments}, however, are often limited in speed of transformation (e.g., slow transformation with inflatables~\cite{suzuki2020lifttiles, teng2019tilepop}) and the range of supported interactions (e.g., only walking~\cite{iwata1999walking}).

This paper introduces \tool{}, a room-scale dynamic haptic environment for virtual reality.
\tool{} provides haptic sensations by reconfiguring physical environments using a small swarm of robot assistants. Inspired by shelf-moving robots~\cite{guizzo2008kiva, wurman2008coordinating} that are used in robotic warehouses, we \changes{developed} a swarm of shape-changing robots that can move a range of existing furniture. Each robot has a mechanical lift that extends from 30 cm to 100 cm to pick up, carry, and place objects such as chairs, tables, and walls. This way, users can touch, sit, place, and lean against objects in the virtual environment. To synchronize the VR scene with the physical environment in a 10 m x 10 m space, we developed software to track and control the robots with an optical motion capture system. This system continuously maps virtual touchable surfaces in the proximity of users and coordinates the robot swarm to move physical objects to their target location without colliding with each other or the users.

We investigate the use of \tool{} for real estate virtual tours and collaborative architectural design, two increasingly common application areas for VR~\cite{ibayashi2015dollhouse}.
\changes{To support these scenarios, we propose four types of basic interactions along with the spectrum between
{\it embodied} and {\it controller-based} interactions: 1) walking and touching, 2) physically moving furniture, 3) teleporting, and 4) virtually moving furniture.
We describe and demonstrate how these interactions can be used for architectural design and virtual tour scenarios, such as exploration of the architectural space, remote collaboration and co-designing, navigation in the large space, and virtual scene editing.}
In a preliminary evaluation with five participants, we test the feasibility of moving furniture robotically to simulate a static physical environment. Based on our insights, we discuss future research directions.

In short, we contribute:
\begin{enumerate}[itemsep=-1mm]
\item A concept of providing room-scale dynamic haptic feedback through furniture-moving swarm robots.
\item Design and implementation of \tool{}: mechanical design of a shape-changing robot with a scissor lift, tracking techniques, software, and hardware implementation to synchronize the physical environment with the virtual scene.
\item Application scenarios for virtual tours and architectural design and appropriate interaction techniques.
\end{enumerate}

%% file: 2-related-work.tex
\section{Related Work}

\subsection{Haptic Interfaces for VR}
\changes{There are many approaches to providing haptic sensations for VR.
Existing approaches can be largely categorized in two ways: using {\it passive} objects for haptic props or using {\it active} objects for dynamic haptic feedback.}

\changes{
\subsubsection{Haptic Feedback using Passive Objects}
The first approach uses existing passive artifacts as a haptic proxy for virtual objects~\cite{insko2001passive}.}
For example, Annexing Reality~\cite{hettiarachchi2016annexing} employs physical static objects as props in an immersive environment by matching and adjusting the shape and size of the virtual object.
\changes{Haptic Retargeting~\cite{azmandian2016haptic} uses visual illusions to simulate multiple virtual objects with a single physical prop.
Similarly, by combining passive objects with redirected walking~\cite{razzaque2005redirected}, Kohli et al.~\cite{kohli2005combining} explored haptics that can go beyond the scale of human hands.}
Using passive objects as haptic proxies benefits from easy and low-cost haptic feedback, but it is difficult to fully represent the dynamic virtual objects because the physical props are static.

\changes{
\subsubsection{Hand-held and Wearable Haptic Interfaces}
On the other hand, active haptic interfaces leverage actuated devices to provide dynamic haptic feedback.
In the literature, the active haptic feedback has been mostly explored as hand-held or wearable for on-body haptic feedback.
For example, prior work explores haptic feedback for hands or finger-tip using a controller with actuated pins~\cite{benko2016normaltouch}, graspable exoskeleton~\cite{choi2017grabity, choi2016wolverine}, and various hand-held controllers that generate force feedback~\cite{heo2018thor, je2019aero,sasaki2018leviopole,shigeyama2019transcalibur}.}
On-body haptic interfaces have been also investigated to provide haptic sensations beyond the human hands~\cite{lindeman2006wearable}.
For example, wearable haptic suits~\cite{nagai2015wearable, delazio2018force}, electric muscle simulation~\cite{lopes2015impacto}, and actuated shoes~\cite{schmidt2015level} have been proposed to provide haptic feedback to the user's body.
Hand-held and wearable approaches have many benefits in portability and mobility, but current hand- or body-based haptic interfaces are still limited to emulate a whole environment, which is the focus of our work.

\changes{
\subsubsection{Encountered-type Haptic Feedback with Robotic Devices}
The other approach of active haptic feedback is {\it encountered-type haptics}, in which haptic devices dynamically deliver physical props when the user makes contact with a virtual object.
Unlike the wearable approach, the encountered-type approach augments the environment, instead of the user's body.}
Ever since McNeely's concept of Robotic Graphics~\cite{mcneely1993robotic}, robots in various form factors have investigated this encountered-type approach to emulate virtual objects and environments.
\changes{For example, existing work uses shape displays to simulate dynamic surface and shapes for VR (e.g., FEELEX~\cite{iwata2001project}, shapeShift~\cite{abtahi2018visuo,siu2018shapeshift}), robotic arm to simulate walls and objects (e.g., Snake Charmer~\cite{araujo2016snake}, VRRobot~\cite{vonach2017vrrobot}), and multiple movable robots to simulates dynamic terrain on which the user walks (e.g., CirculaFloor~\cite{iwata1999walking}).
Recently, drones~\cite{Abtahi:2019:BFU:3290605.3300589, hoppe2018vrhapticdrones} and small wheeled robots~\cite{he2017physhare} have been also proposed as a way to move proxy objects for encountered-type haptic experiences.
Our work is also categorized as the encountered-type haptics, but in contrast to existing approaches, we aim to emulate the whole environment of the virtual space. 
We achieve this by combining passive objects with robotic actuation. 
Instead of emulating building elements with the robot itself, we propose actuating furniture with robots, which allows more robust and larger-scale haptic feedback for VR.}

\subsubsection{Large-scale Haptic Feedback}
\changes{Prior work also explores large-scale haptic feedback by reconfiguring the physical environment around the user.}
HapticTurk~\cite{cheng2014haptic} aims to achieve this goal through the human actuation method, in which human helpers lift and tilt a player to produce a physical sensation in a video game. Similarly, TurkDeck~\cite{cheng2015turkdeck} leverages human volunteers to help to reconstruct the space for realistic haptic feedback. 
The human actuation approach~\cite{cheng2018iturk, cheng2014haptic,cheng2017mutual,cheng2015turkdeck} is intriguing for prototyping or simulating the large-scale haptic feedback, but is limited in availability and scalability as it always requires a group of human helpers to achieve the experience.
In contrast, room-scale systems such as TilePoP~\cite{teng2019tilepop} and LiftTiles~\cite{suzuki2020lifttiles} investigate the use of inflatable actuators to dynamically change the environment to provide a haptic proxy.
Although inflatables are safe and low-cost, they are also slow: the speed of transformation and degrees of freedom are key limitations to providing real-time dynamic haptic sensations.
\changes{To overcome this limitation, this paper explores a swarm robotic approach to reconfigure the environment. This benefits from faster spatial transformation which is important for the real-time synchronization between virtual and physical environments.}

\subsection{Human-Robot Interaction with Swarm Robots}
Outside of the context in haptics for VR, our work is also inspired by an emerging area of human-robot interaction, particularly with swarm robots~\cite{le2016zooids, kim2017ubiswarm}.
Some existing works explore the use of these small tabletop robots for everyday haptic feedback~\cite{guinness2018haptic,kim2019swarmhaptics, suzuki2017fluxmarker} or constructing haptic proxy objects for VR~\cite{zhao2017robotic}.
We extend these works by exploring the large-scale haptic feedback, inspired by the concept of shape-changing swarm robots~\cite{suzuki2019shapebots}.
In the context of human-robot interaction, some works integrate wheeled robots with furniture for spatial reconfiguration.
For example, Mechanical Ottoman~\cite{sirkin2015mechanical} adds a wheeled robot to existing furniture to make it autonomously mobile. Similarly, shape-shifting wall displays move to adapt to the content projected on the wall~\cite{takashima2016study}.
We extend prior work on robotic furniture by investigating its use for haptics in VR and proposing the required technologies and interaction techniques unique to VR applications.




%% file: 4-system.tex
\section{RoomShift: System and Implementation}
\subsection{Overview}
\tool{} consists of a small swarm of shape-changing robots; each robot uses a Roomba as a mobile base. On this base is mounted a custom mechanical scissor lift made of two linear actuators and a metal drying rack.
As the mechanical lift is compact in its closed state, the robot can move under a table or chair with 30 cm clearance, and extend the scissor lift to pick it up.

\begin{figure}[h!]
\centering
\includegraphics[width=1\columnwidth]{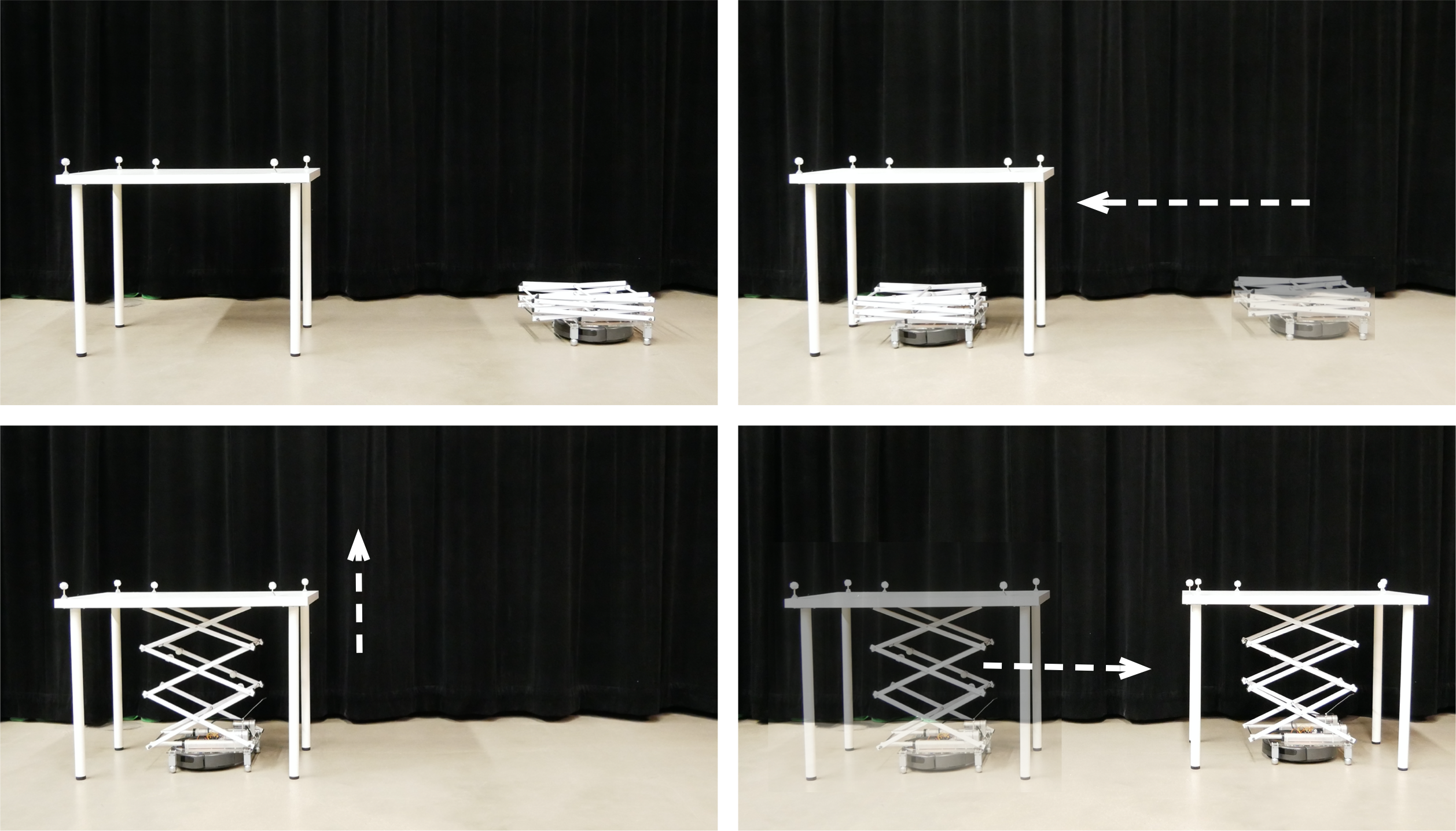}
\caption{A RoomShift robot drives beneath a desk, lifts it by extending the scissor structure, and moves it.}
~\label{fig:system-mechanism}
\vspace{-0.4cm}
\end{figure}

The goal of \tool{} is to provide whole-body haptic interactions in VR.
It does this by dynamically constructing and reconfiguring physical spatial layouts.
Our approach is to reconfigure the physical room using a swarm of furniture-moving robots.
These robots can relocate existing physical elements of the environment (e.g., chairs, racks, shelves, desks, as well as custom props) \changes{as encountered-type haptic interfaces that can support whole-body interactions.} 
The key inspiration comes from the shelf-moving robots~\cite{guizzo2008kiva,wurman2008coordinating} that are used in robotic warehouses. By bringing similar capabilities to virtual reality, we can achieve fast, robust, large-scale, and scalable dynamic haptic environments.

The position and orientation of the robots and physical props are tracked with an optical motion capture system through five retro-reflective markers, \changes{whose patterns are unique to be identified as different objects}.
The motion capture system also tracks the position of the user's VR head-mounted display, so the user can walk around within a 10 m $\times$ 10 m area.
A VR scene is rendered through the A-Frame API on an Oculus Go headset and the position information of the VR scene is synchronized through a server between the VR headset and the desktop computer that tracks objects and controls the robot swarm.
Whenever the user loads a new scene, teleports to a new location, or changes the design of the virtual layout, the system calculates the appropriate position of physical props and drives the robots to reconstruct the physical environment.

\begin{figure}[h!]
\centering
\includegraphics[height=4.05cm]{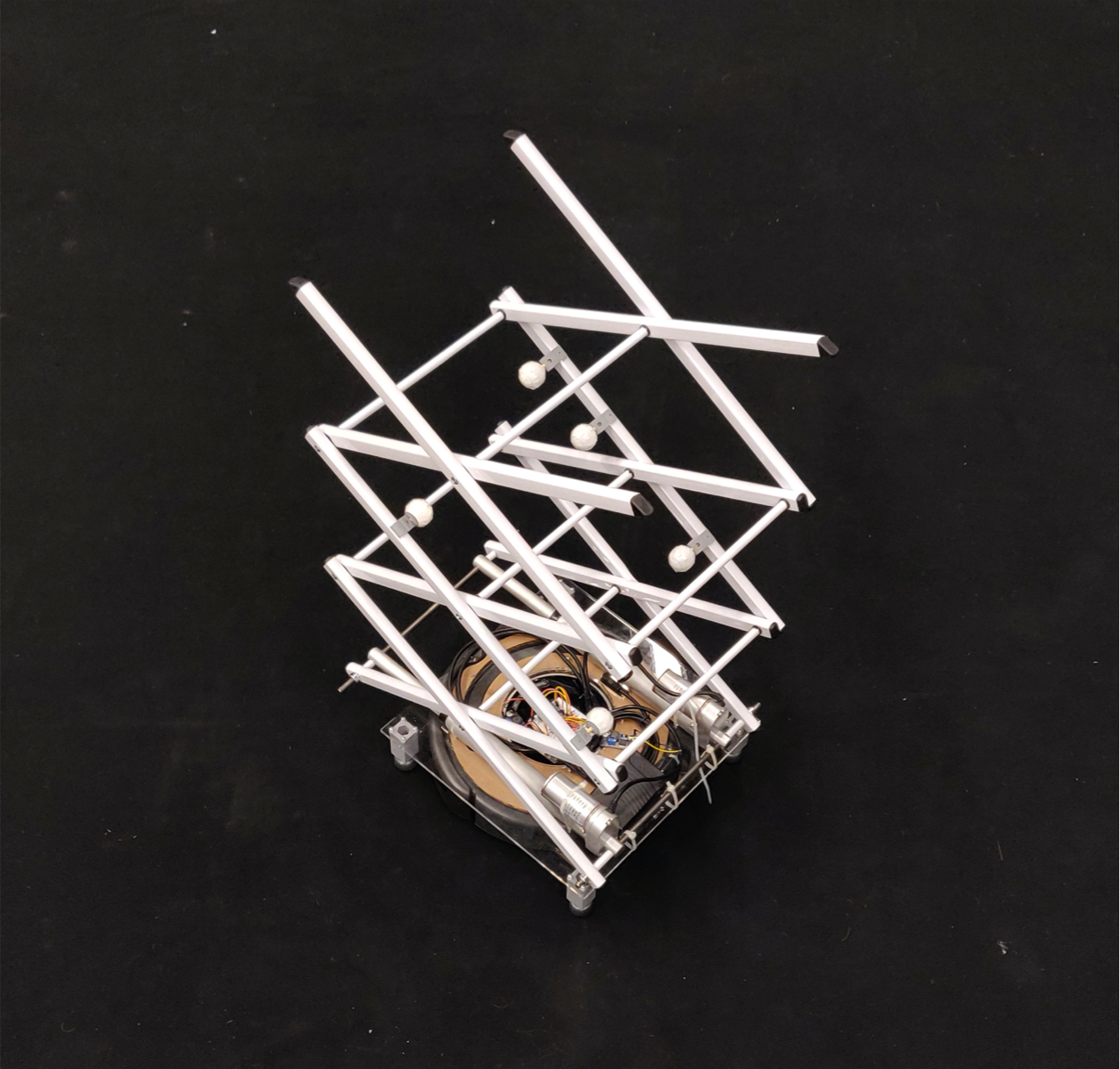}
\includegraphics[height=4.05cm]{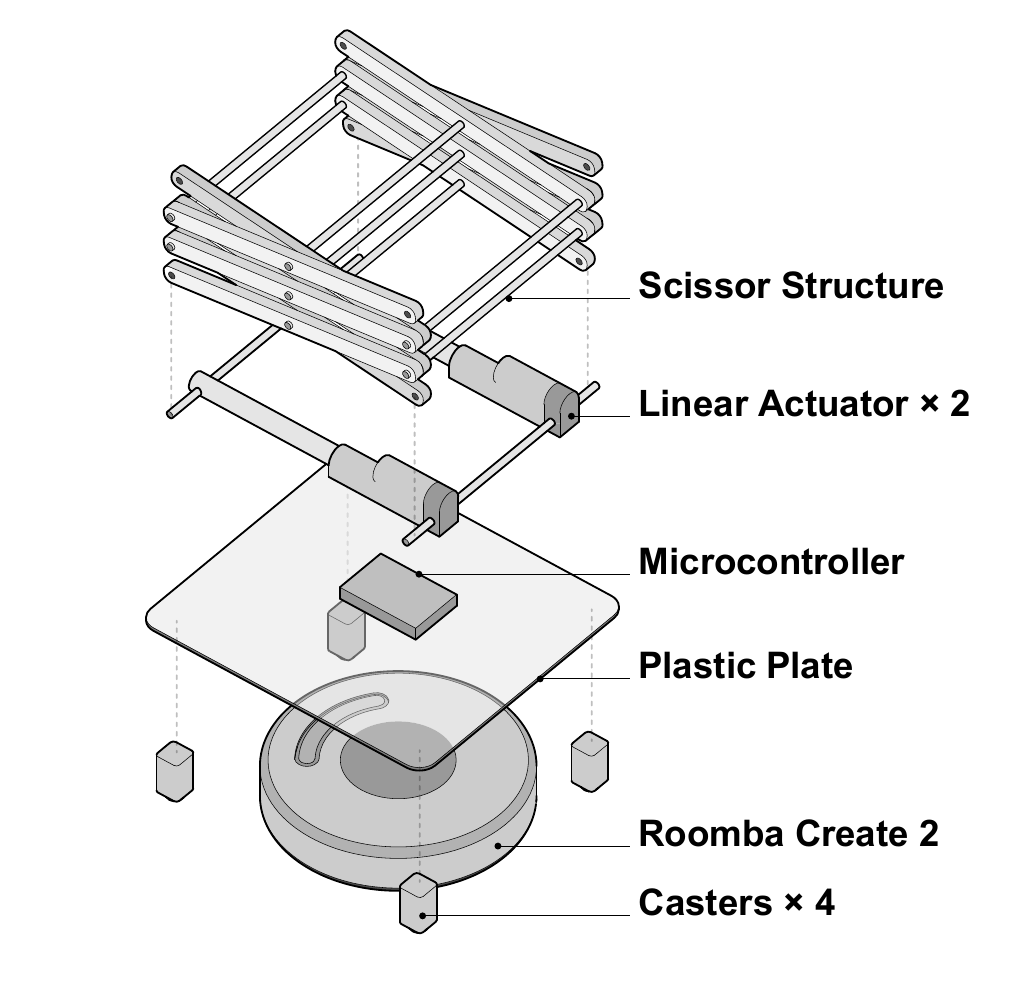}
\caption{Mechanical design of the robot and the scissor structure.}
~\label{fig:system-mechanical-design}
\vspace{-0.4cm}
\end{figure}

\subsection{Mechanical Design}
\tool{} comprises nine shape-changing swarm robots based on the Roomba Create 2~\cite{dekan2013irobot}.
For the mechanical lift structure, we repurposed an off-the-shelf expandable laundry rack (Room Essentials Compact Drying Rack) and attached two linear actuators (Homend DC12V 8 inch Stroke Linear Actuator, which extends from 32 cm to 52 cm) at the base of the rack.
The linear actuators are fixed to the endpoints of the scissor structure with 8 mm steel rods, so that when the actuator contracts, the mounted scissor structure extends vertically (from 30 cm to 100 cm).
The scissor structure moves at a speed of 1.3 cm / sec. To mount the scissor structure, we fixed a 6mm acrylic bottom plate (35 cm x 35 cm) and four omni-directional casters (Dorhea Ball Transfer Bearing Unit) to relieve the Roomba of most of the weight that the robot carries. Each robot moves at 20 cm / sec.
Figure~\ref{fig:system-mechanical-design} illustrates the mechanical design of each \tool{} robot.


\begin{figure}[h!]
\centering
\includegraphics[width=1\columnwidth]{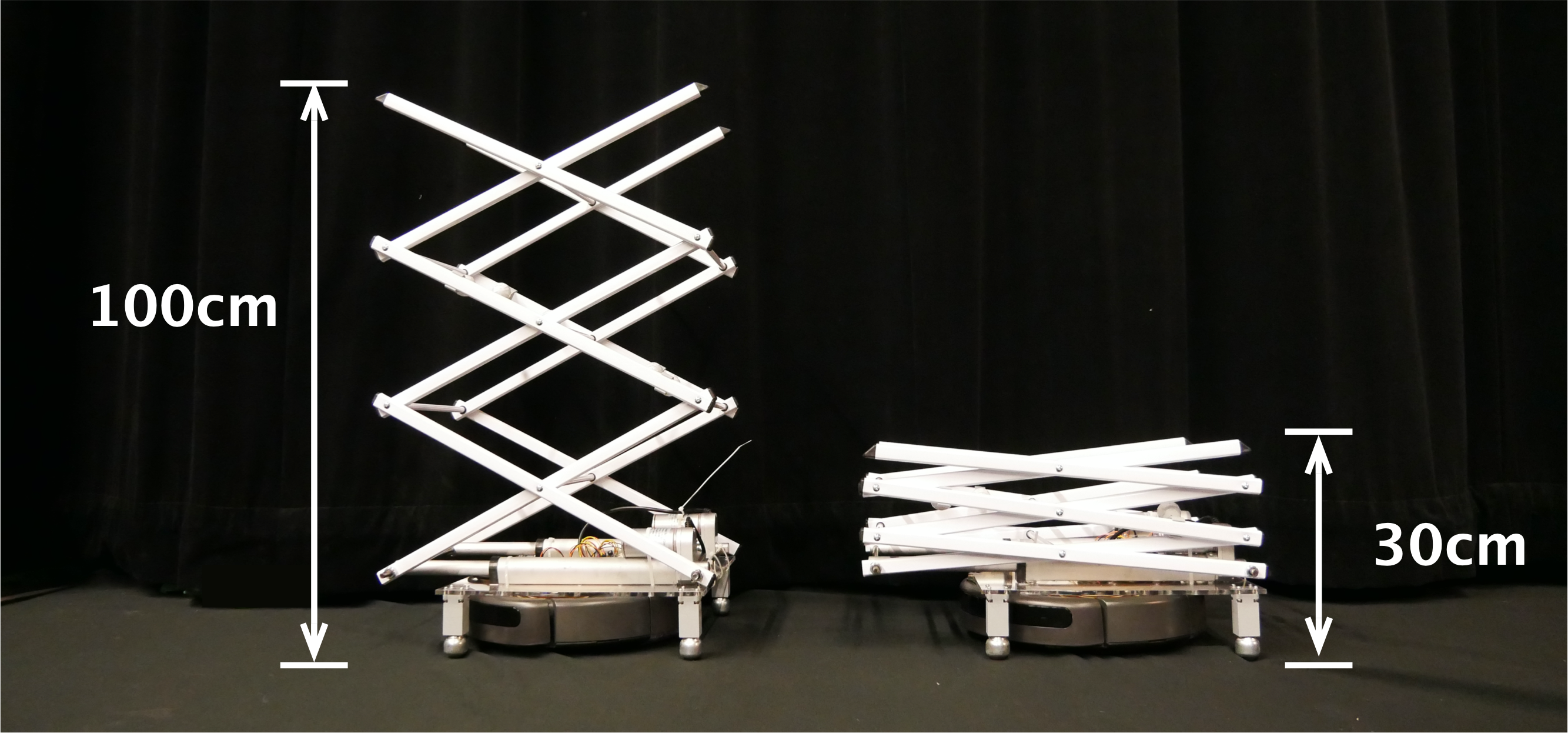}
\caption{Each robot can extend from 30 cm to 100 cm to lift objects.}
~\label{fig:system-structure}
\vspace{-0.4cm}
\end{figure}

We considered and tested several actuation mechanisms such as a pneumatically-actuated inflatable structure~\cite{hammond2017pneumatic, suzuki2020lifttiles, teng2019tilepop}, a deployable structure using coilable masts~\cite{jensen2001arm, joosten2007preliminary}, and a mechanical reel-based actuation~\cite{takei2011kinereels}.
Pneumatic actuation is problematic for our mobile setup as it requires a tube connected to a pump or pressure tank to supply air. The deployable structure and mechanical reel-based actuation affords a much higher extension ratio, but is limited in its robustness and load-bearing capability. The mechanical scissor structure is appropriate for our purpose because it is inexpensive (compact drying rack: \$ 15, linear actuators: \$ 35 x 2) and lightweight (2kg).
Existing warehouse robots such as Kiva~\cite{guizzo2008kiva} have a limited expandable capability as they are designed for one specific shelf, whereas our mechanical scissor lift can move various objects by leveraging its highly expandable structure (4 times expansion ratio). 
\changes{The current actuator height (30 - 100 cm) was chosen to cover a wide range of standard chairs and tables, which measure 30 - 76cm and 48 - 96 cm respectively~\cite{standardfurniture}.}
The maximum height of the scissor structure itself can be also extended by adding more elements \changes{like combining two scissor structures to double the maximum height, with the trade-off with the less stable structure.}


\subsection{Object Actuation}
\changes{One advantage of our approach} is that the robot need not support the weight of the user.
\changes{Once the robot places the furniture, it serves as a static object. Thus, when a user sits on or puts weight on it, all of the weight goes to the furniture, instead of the robot}, which significantly reduces the possibility of a mechanical breakdown.

\begin{figure}[h!]
\centering
\includegraphics[width=1\columnwidth]{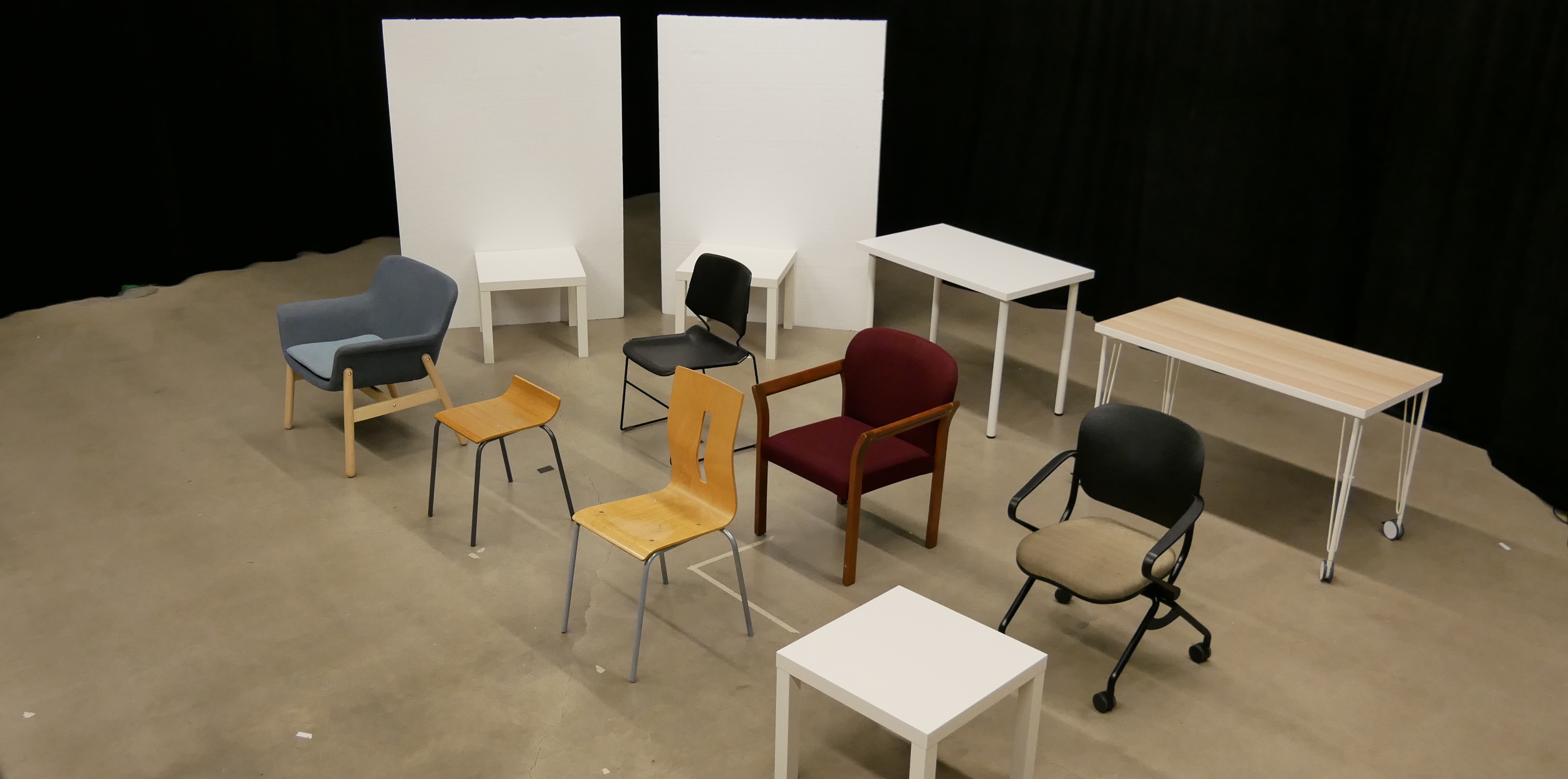}
\caption{Different types of furniture moved by the system.}
~\label{fig:system-object-types}
\vspace{-0.4cm}
\end{figure}

Although the maximum load for the Roomba is 9 kg, the corner-mounted casters distribute and carry heavier loads.
Thus, our robots can lift and carry heavier objects than an unmodified Roomba.
\changes{The maximum weight the robot can lift and carry is 22 kg. When we put a heavier object than 23 kg, we observed the scissor structure started to break.
The strength of the scissor structure suffices to lift lightweight chairs and tables, such as the IKEA honeycomb furniture used in our prototypes.
The weight of the furniture we have tested (depicted in Figure~\ref{fig:system-object-types}) ranges from 3.5 to 11.2 kg.}
For heavier objects, multiple robots can also coordinate to lift a piece together if there is sufficient space under the furniture.
Also, with a more robust scissor structure, we can carry heavier objects, as we observed the Roomba base itself (with the corner-mounted casters) can carry up to 30 kg load.

\changes{This approach also increases flexibility because different types of furniture can be actuated with the height-adjustable scissor lift.}
For example, Figure~\ref{fig:system-object-types} illustrates various static props that the \tool{} robot can actuate.
These objects include furniture such as a desk, a long table, different chairs, and a side table.
Note that due to the robot's minimum collapsed size, objects must have at least 30 cm clearance below them, and enough horizontal space to fit the robot. A designer can also create custom props for specific applications, for instance, the styrofoam wall mounted to a side table seen in Figure~\ref{fig:system-object-types}.

\subsection{Tracking System}
\begin{figure}[h!]
\centering
\includegraphics[width=1\columnwidth]{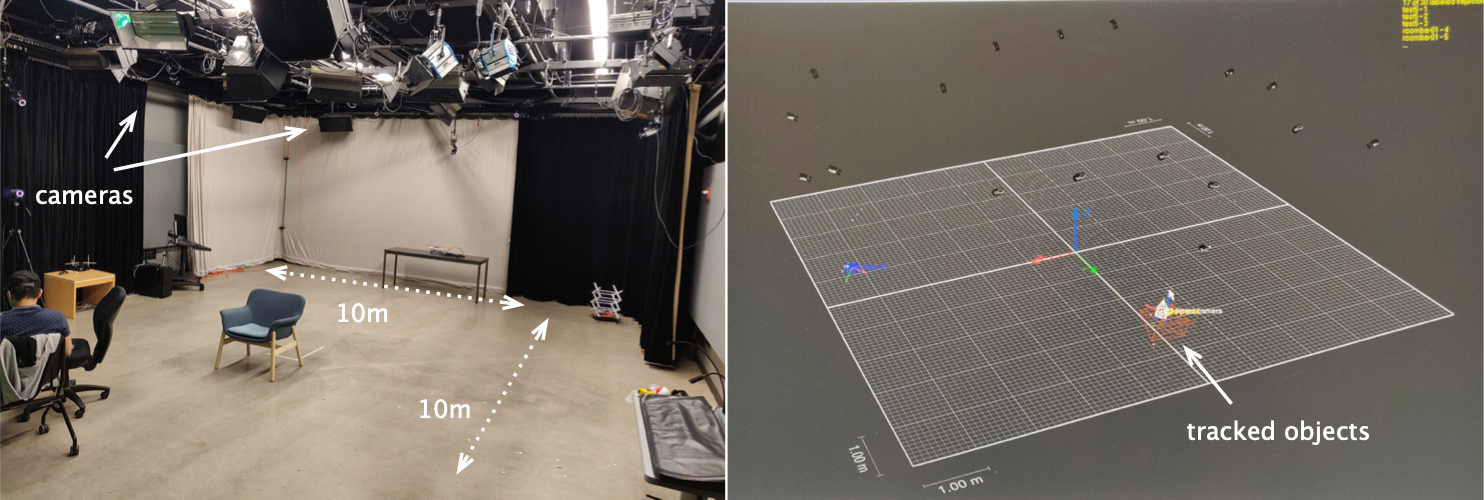}
\caption{Photo of tracked space and screenshot of tracking software.}
~\label{fig:system-tracking}
\vspace{-0.4cm}
\end{figure}

To accurately control the \tool{} robots, we require precise motion tracking that can cover the play area in which a user walks. We use an optical tracking system with 20 IR cameras (Qualisys Miqus 5) that can track objects in a 10 m $\times$ 10 m space. Figure~\ref{fig:system-tracking} depicts the space and mounted cameras on the ceiling (left) and tracking software (right). The system tracks six degrees of freedom (DOF) position of the objects with retro-reflective spherical markers at 60 FPS frame rate.

\begin{figure}[h!]
\centering
\includegraphics[width=1\columnwidth]{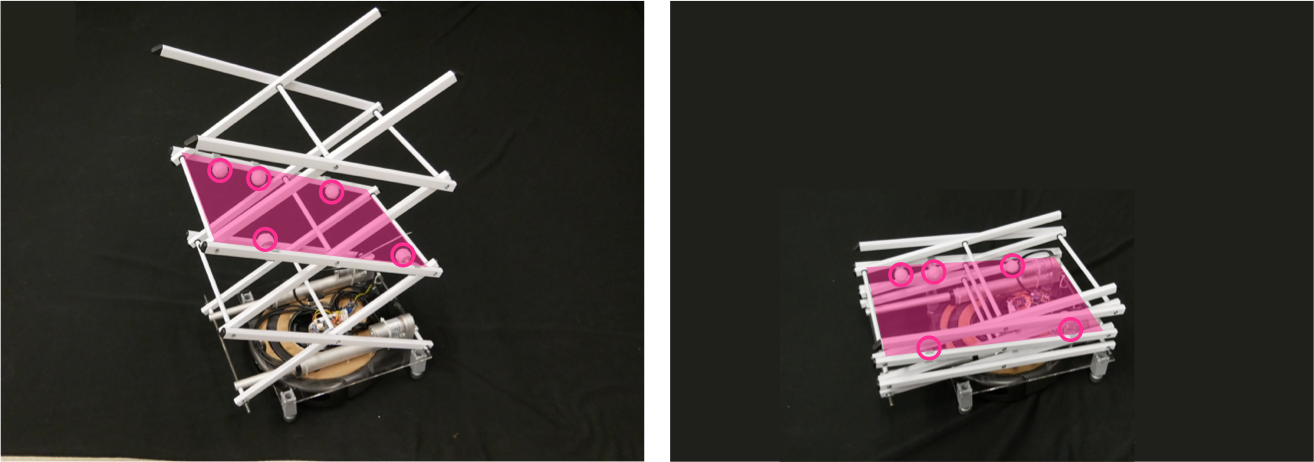}
\caption{Retro-reflective markers mounted to parallel lift bars, highlighted in pink.}
~\label{fig:system-tracking-marker}
\vspace{-0.4cm}
\end{figure}

To track each robot, we attached five 30 mm spherical retro-reflective markers to the bars of the scissor structure (Figure~\ref{fig:system-tracking-marker}). 
We attached markers to a pair of parallel bars, so that the markers' relative positions remain constant regardless of the height of the scissor lift.
We can also estimate the height of the scissor structure by measuring the orientation of the marker pattern (the pink plane surface depicted in Figure~\ref{fig:system-tracking-marker}).

All physical props have retro-reflective markers attached, so that the system can capture and track their positions and orientations, and plan the paths for the robots to pick them up and avoid collisions. They also enable the system to track the robots while moving objects: when markers attached to the robot are hidden beneath an object it is carrying, the system can still reliably track the robot using the object markers as a proxy for the hidden robot.
\changes{Our motion capture system uses 20 cameras, so even when large objects (e.g., walls) occlude furniture trackers from the cameras in one direction, opposing cameras maintain tracking. We tested with multiple walls, and when the furniture is enclosed from two sides, the system starts to lose tracking. However, in testing, we did not encounter this scenario frequently.}

\subsection{Control System}
To control the robots' movements, we use a simple path planning algorithm. 
The input is 1) the current positions of the robots, 2) the positions of obstacles (e.g., furniture, other robots, and users), and 3) the target locations.
The algorithm outputs the goal of each robot at the next time step.
The system continuously \changes{updates the path} and drives them to their target locations. The main server continuously tracks the robot positions, calculates their wheel speeds, and sends commands at 30 Hz over WiFi.

\changes{Here, we describe the control algorithm in more detail.}
The first step of the algorithm is to assign the target position for each robot.
The system assigns the optimal target for each robot by solving an assignment problem.
We first constructs a distance matrix $D = (d_{i, j})$ where $d_{i, j}$ represents the distance between the robot $i$ and target $j$.
The system computes the optimal combination by applying the Hungarian algorithm to the distance matrix.
Given the assigned target, the system computes the path based on simple Reciprocal Velocity Obstacles (RVO) algorithm~\cite{van2008reciprocal}.
This algorithm also handles collision avoidance with users, other robots, and obstacles. 

\begin{figure}[h!]
\centering
\includegraphics[width=1\columnwidth]{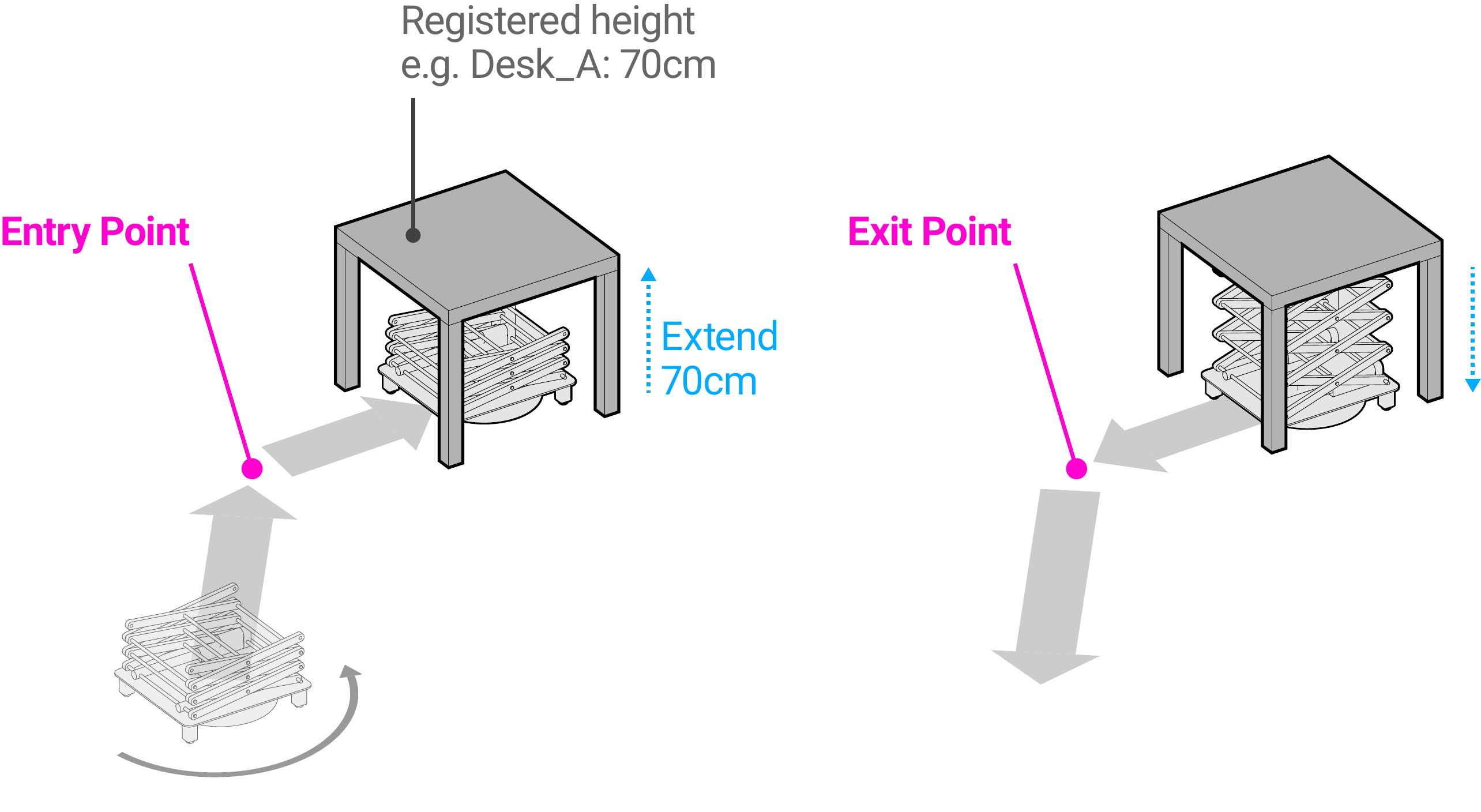}
\caption{\changes{The system first navigates the robot to a user-defined entry point to avoid the collision with the legs of furniture.}}
~\label{fig:system-control}
\vspace{-0.4cm}
\end{figure}

\changes{With the RVO algorithm, we get the vector the robot should move at each time step. Based on this, the control system determines the speed of the left and right wheels (ranged from -255 to 255) with the PID control.
The error function of the PID control is $e(t) = da_t$, where $da_t$ is the difference between the current orientation and the target vector. The system moves the robot by minimizing this $e(t)$ with the standard PID control function which is $u(t) = K_p e(t) + K_i \int_{0}^{t}e(t')dt' + K_d de(t)/dt$, where $K_x$ is the gain parameter of each factor.
We determine the speed of left and right wheels as $A(dd_t) * (1 + u(t))$ and $A(dd_t) * (1 - u(t))$ respectively. $A(dd_t)$ is the linear function of the difference in distance at time $t$, with a certain minimum and maximum threshold, so that when the distance is smaller than a certain threshold, the robot stops because $A(dd_t)$ becomes zero.}


To pick up and place these, the robot follows a predefined sequence, approaching the object from an angle where it will not collide with the object's legs.
To avoid the collision with the legs of furniture, each object has a user-defined entry and exit point (Figure~\ref{fig:system-control}).
\changes{We also register the height of target furniture before the system starts (e.g., 70 cm for Table\_A, 40 cm for Chair\_B), so that it can extend the scissor lift to certain target height. 
We could also put a simple sensor on top of the scissor structure to make it a closed-loop system.}

\subsection{\changes{Electronics Implementation}}
\begin{figure}[h!]
\centering
\includegraphics[width=1\columnwidth]{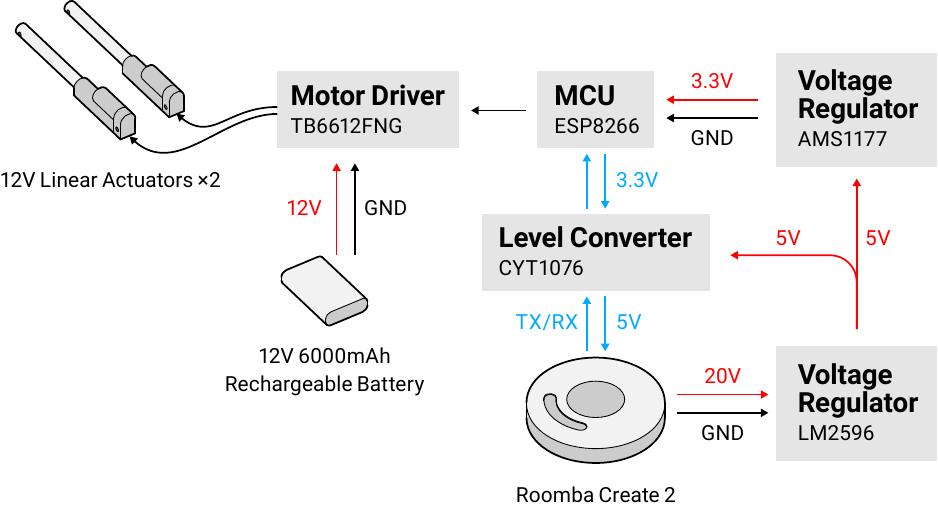}
\caption{Hardware schematic of the robot.}
~\label{fig:system-schematics}
\vspace{-0.4cm}
\end{figure}

Figure~\ref{fig:system-schematics} illustrates the schematic of \tool{}'s circuit. 
Each robot is controlled with an ESP8266 microcontroller chip (Wemos D1 mini), powered by the Roomba through a voltage regulator. 

The power source of the microcontroller is the Roomba's internal battery which supplies 14-20V. A voltage regulator (LM2596) first steps this down to 5V. The 5V power is supplied to another voltage regulator (AMS1177), which supplies 3.3V to the ESP8266 microcontroller and logic level converter (CYT1076). The logic level converter converts the voltage for serial communication between the microcontroller (3.3V) and Roomba (5V). The microcontroller receives commands over WiFi and controls the left and right wheels of the Roomba using a PWM signal. The microcontroller also operates two linear actuators using a dual motor driver (TB6612FNG). The Roomba's internal battery is insufficient to supply the current for the linear actuators (600-800mA for average, 1.5-2A for peak current), so we use an external portable rechargeable battery (12V 6000mAh) to power the actuators.

\begin{figure}[h!]
\centering
\includegraphics[width=1\columnwidth]{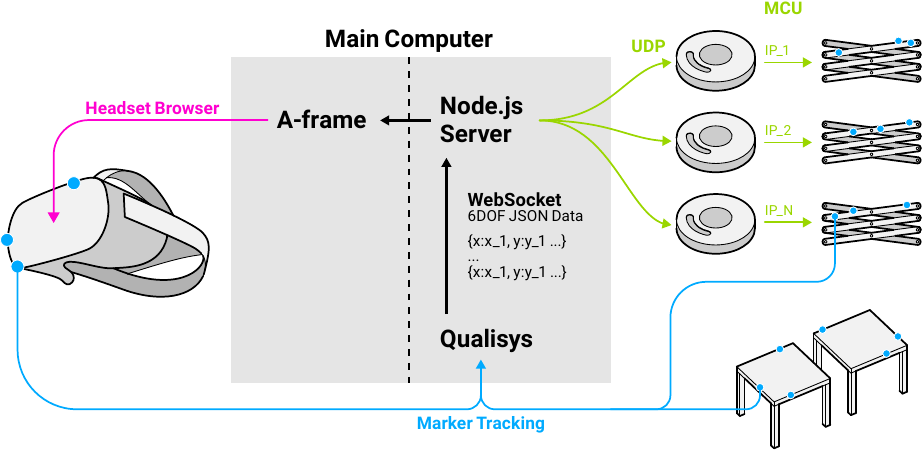}
\caption{The communication software.}
~\label{fig:system-block-diagram}
\vspace{-0.4cm}
\end{figure}

\begin{figure*}[t!]
\centering
\includegraphics[width=1\textwidth]{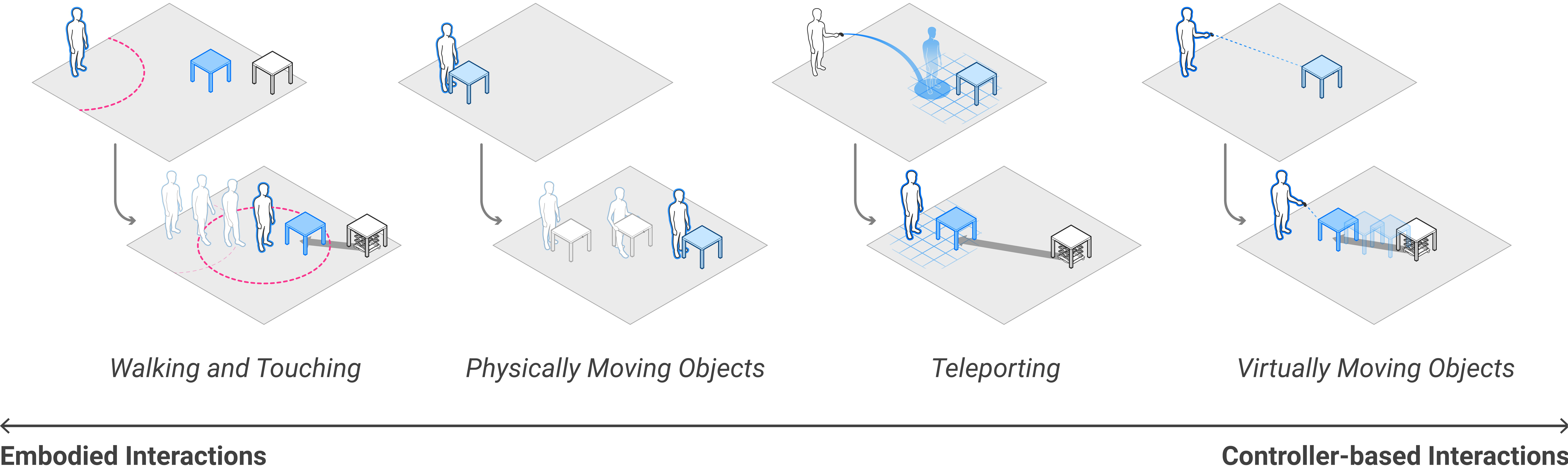}
\caption{\changes{The interaction design space of \tool{}.}}
~\label{fig:interaction-design-space}
\vspace{-0.4cm}
\end{figure*}

\subsection{Software Implementation}
Figure~\ref{fig:system-block-diagram} illustrates the architecture of the \tool{} software. 
The main computer runs a Node.js server and the Qualisys tracking software. 
The 6DOF tracking data that the Qualisys tracking system captures is streamed to the Node.js server through the WebSocket protocol. 

Based on the tracking data, a web browser client renders the VR scene with A-Frame.
The user experiences the VR scene using an Oculus Go head mounted display and its built-in VR browser.
We synchronize the desktop computer and the Oculus Go browser with real-time communication through WebSocket.
When the virtual scene changes, the system moves the robots to dynamically reconfigure the physical scene.
First, the system computes the types of props and each target position based on the relative position from the user.
Once the target position of the physical props is determined, the Node.js server sends commands to each robot over WiFi. 

\subsection{Preliminary System Evaluation}
We conducted a preliminary evaluation to understand how the system would function and gauge user experiences in VR enabled by \tool{}. We posit that the system can reconfigure a simple virtual scene to the point where participants' haptic experiences match those of a static physical environment, and that limitations like robot speed, and physical movement would not diminish the sense of immersion.

We recruited five participants (male: 4, female: 1; between 23--34 years old) from our local institution. All participants had prior experience with VR. In a within-subjects counterbalanced design, participants interacted with physical chairs in a VR scene in two conditions: 1) with physical chairs moved by robots and 2) with static physical chairs. We rendered a virtual 7 m $\times$ 7 m room with 4 chairs in VR, matched by a physical area of the same size. \changes{We kept a 1.5 m margin around the play area for additional space that robots can move in or stay.}
Participants wore a head-mounted display (Oculus Go) and noise canceling headphones playing white noise. They were asked to locate a highlighted red chair in the VR scene and sit on it. When sitting, another chair in the room would turn red and the participant was then asked to walk over and sit on the new highlighted chair. This task was repeated eight times in two different conditions. In the first condition, the physical play area contained 4 static physical chairs, which matched the positions of the virtual chairs. In the second condition, RoomShift robots moved two chairs to simulate four chairs. While the user was sitting on a highlighted red chair, a robot moved another chair to the next target location. After the experiment, we asked participants which condition was more realistic (or if they felt the same), and to provide qualitative feedback on their experiences.

\subsubsection{Results and Feedback}
All participants answered that they perceived the realism of the two conditions as the same. This indicates that \tool{} can simulate an environment as intended. While three participants noted they noticed the noise of the moving robots through the headphones, they stated that it did not distract from the realism of the scene. The program did not render a representation of the user's body in VR, and four participants reported that this diminished the perceived realism and their confidence in touching the chairs and sitting down rapidly in both conditions.

Participants provided overall positive feedback in their qualitative responses to \tool{}. 
Participants were particularly enthusiastic about possibilities for Virtual Tourism and Architectural Design:
{\it ``Familiarizing yourself with a place you haven't had a chance to be yet. Maybe someone that is wary of new places, like an agoraphobe, could test out somewhere new as a sort of practice before going there and getting outside of their comfort zone. (P4)''} Users also saw potential to use \tool{} for architectural design: {\it ``Having a model of a proposed space with the ability to interact with furniture (P1).''}
Based on the feedback, we explore more in-depth interactions, specifically focusing on these scenarios.




%% file: 5-applications.tex
\section{Interaction with RoomShift}

\subsection{Target Application Domain}
In this paper, we specifically focus on architectural application scenarios, such as rendering physical room interiors for virtual real estate tours and collaborative architectural design, two increasingly common application areas for VR~\cite{ibayashi2015dollhouse}.
Virtual real estate tours reduce the time and cost compared to on-site viewings, but currently lack the bodily experience of being able to touch surfaces and sit down. In architectural design, VR aids the communication between architects and clients, where proposed designs can be experienced, discussed and modified before building them. We are motivated by how \tool{} can enable people with various physical abilities to experience, test and co-design these environments with their bodies.
Most of the elements in these applications can be covered with a finite set of furniture and props (e.g., chairs, desks, and walls). We discuss some of the basic interactions to support these applications.

\changes{
\subsection{Interaction Design Space}
To support these scenarios, we propose four types of basic interactions \tool{} can support, with the spectrum between embodied interactions and controller-based interactions, as illustrated in Figure~\ref{fig:interaction-design-space}. Embodied interactions refer to interaction with virtual scenes through physical movements and manipulation.
The user can implicitly interact with the system by walking around or explicitly interact with the virtual scene by physically moving furniture.
On the other hand, the user can also interact with the virtual scene with controller-based gestural interactions.
An example is when the user relocates a distant piece of furniture or remove the wall in the room.
The user can also virtually teleport their location to navigate through space.
We describe each interaction technique in the context of architectural design and virtual real estate tours. 
}

\subsubsection{\changes{Experiencing Architectural Spaces: Walking and Touching}}
\changes{The most basic interaction is to render an architectural space that the user can walk around in and touch. To render the haptic proxies for a large space would require a large number of physical props and robots.
On the other hand, the user's immediate physical reach is usually smaller than the entire virtual scene (e.g., 1.5 m radius).
Therefore, the system only places haptic props within the user's immediate proximity. 
As the user walks around the space, the robots move the props to maintain the illusion of a larger number of objects.
In this way, a small number of robots with a finite set of physical props can suffice to provide haptics for the scene as the system does not need to physically render the entire environment (Figure~\ref{fig:cover}).

\begin{figure}[h!]
\centering
\includegraphics[width=1\columnwidth]{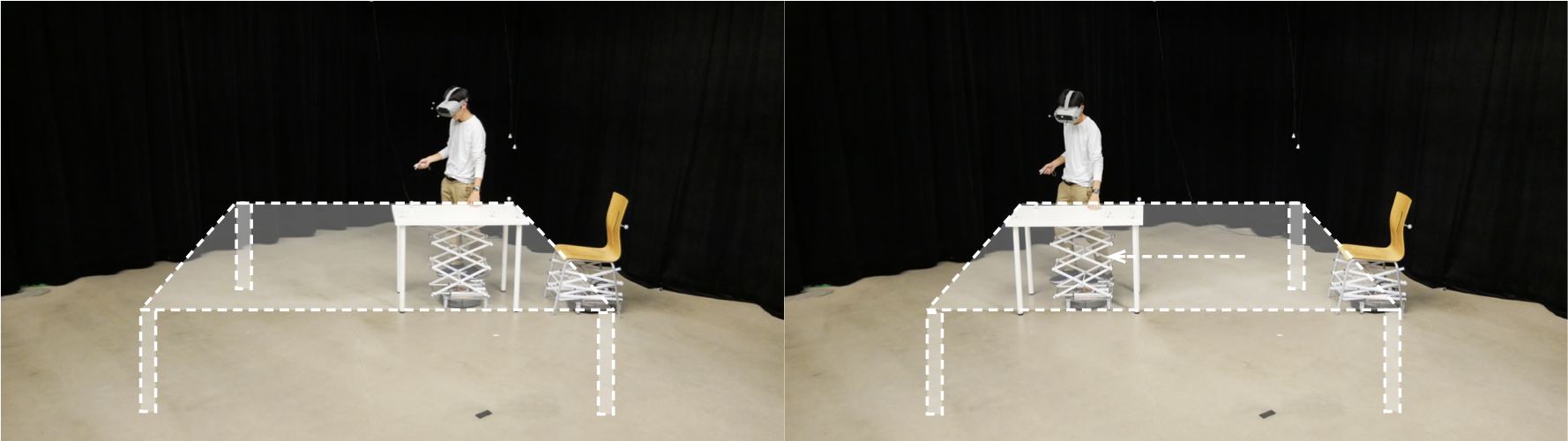}
\caption{Simulating a larger table by moving a smaller surface.}
~\label{fig:interaction-extend}
\vspace{-0.4cm}
\end{figure}

In addition, the system can mimic larger objects with a single moving robot.
For example, when the user is interacting with a large table, either new physical table segments can be added or a single robot can continually move the current table according to the user's position to simulate touching a larger one. This way, a limited number of robots and furniture can simulate large objects (Figure~\ref{fig:interaction-extend}). We also employ this technique for rendering larger wall segments, where the robot moves, carrying the proxy, as the user walks along the wall, similar to a technique proposed in PhyShare~\cite{he2017physhare}.
}

\subsubsection{\changes{Architectural Co-Design: Physically Moving Furniture}}
VR can support teams of architects, designers and their clients to experience and discuss architectural and interior designs.
For example, Dollhouse VR~\cite{ibayashi2015dollhouse} proposes such a possibility for the collaborative design of the home and office spaces, where a user experiences space in VR, while a designer views the layout on a desktop computer remotely and changes the design during the discussion. \tool{} system improves the immersion of this collaborative design process by enabling whole-body interactions with furniture. 
Suppose a situation where a designer and a client are remotely co-designing a new office space at two separated \tool{} systems.
When the designer reconfigures the furniture or space, the robots in the remote location can move the physical objects in real-time to render the designer's change. 
In this way, two remote physical environments can synchronize with a single virtual space.
This aids co-design where the client can touch, feel, walk around, and modify the design in VR.

\begin{figure}[h!]
\centering
\includegraphics[width=1\columnwidth]{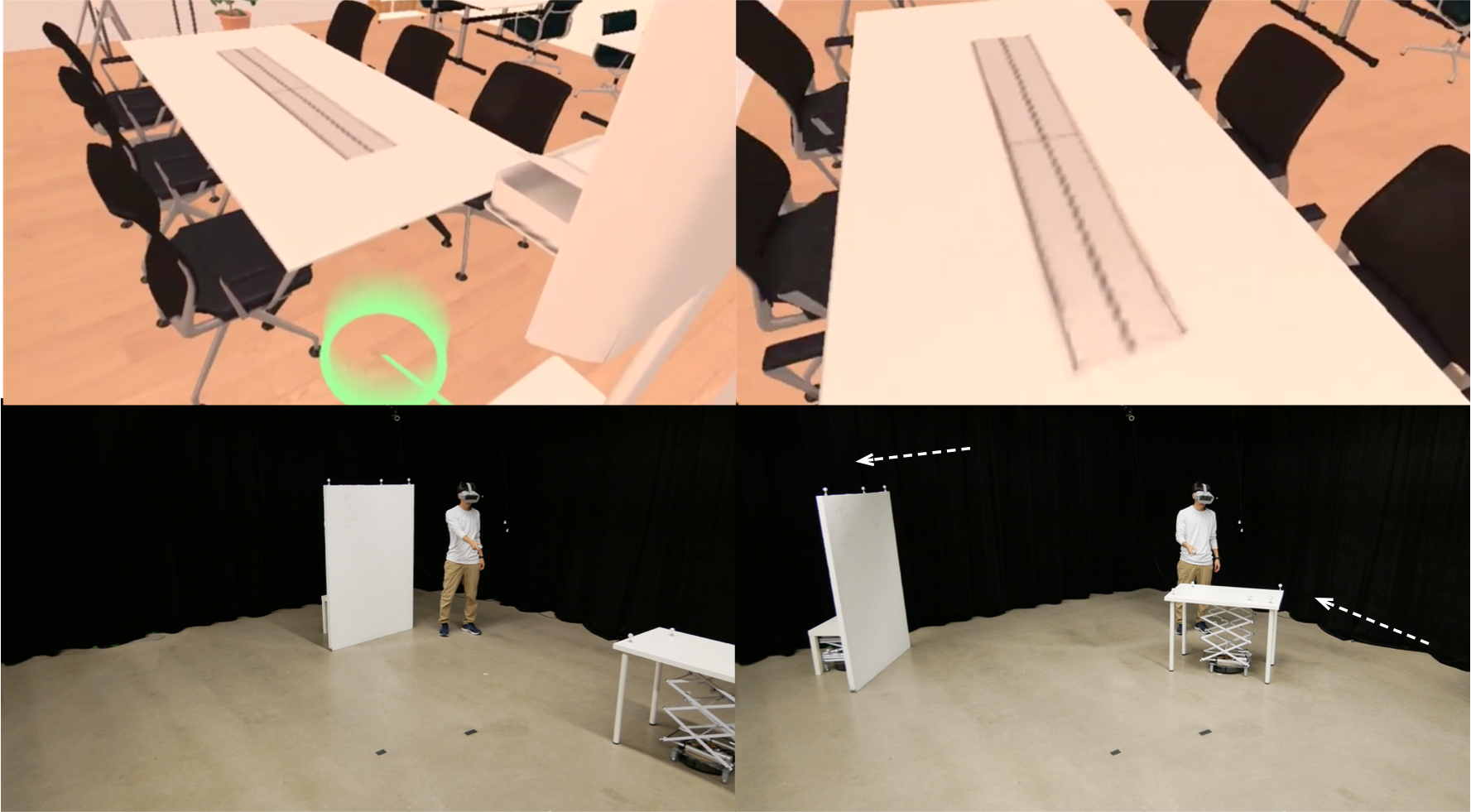}
\caption{When teleporting, the robots move furniture to match the new scene position.}
~\label{fig:interaction-teleport}
\vspace{-0.4cm}
\end{figure}

\subsubsection{\changes{Navigating Large Spaces: Teleporting in VR}}
The physical play area of a VR setup is often much smaller than the virtual scene. Teleportation is a common navigation technique that enables the user to point with a controller to a distant location in the scene and instantly move there \cite{langbehn2018evaluation}.
\tool{} supports teleportation by reconfiguring the room layout to match the new view location (Figure~\ref{fig:interaction-teleport}).
When the user teleports to a new location in the VR scene, the system calculates the positions of the virtual objects relative to the new location and moves the furniture and robots in and out of the play area to enable a fast scene reconfiguration and to avoid collisions with the user and each other.

\begin{figure}[h!]
\centering
\includegraphics[width=1\columnwidth]{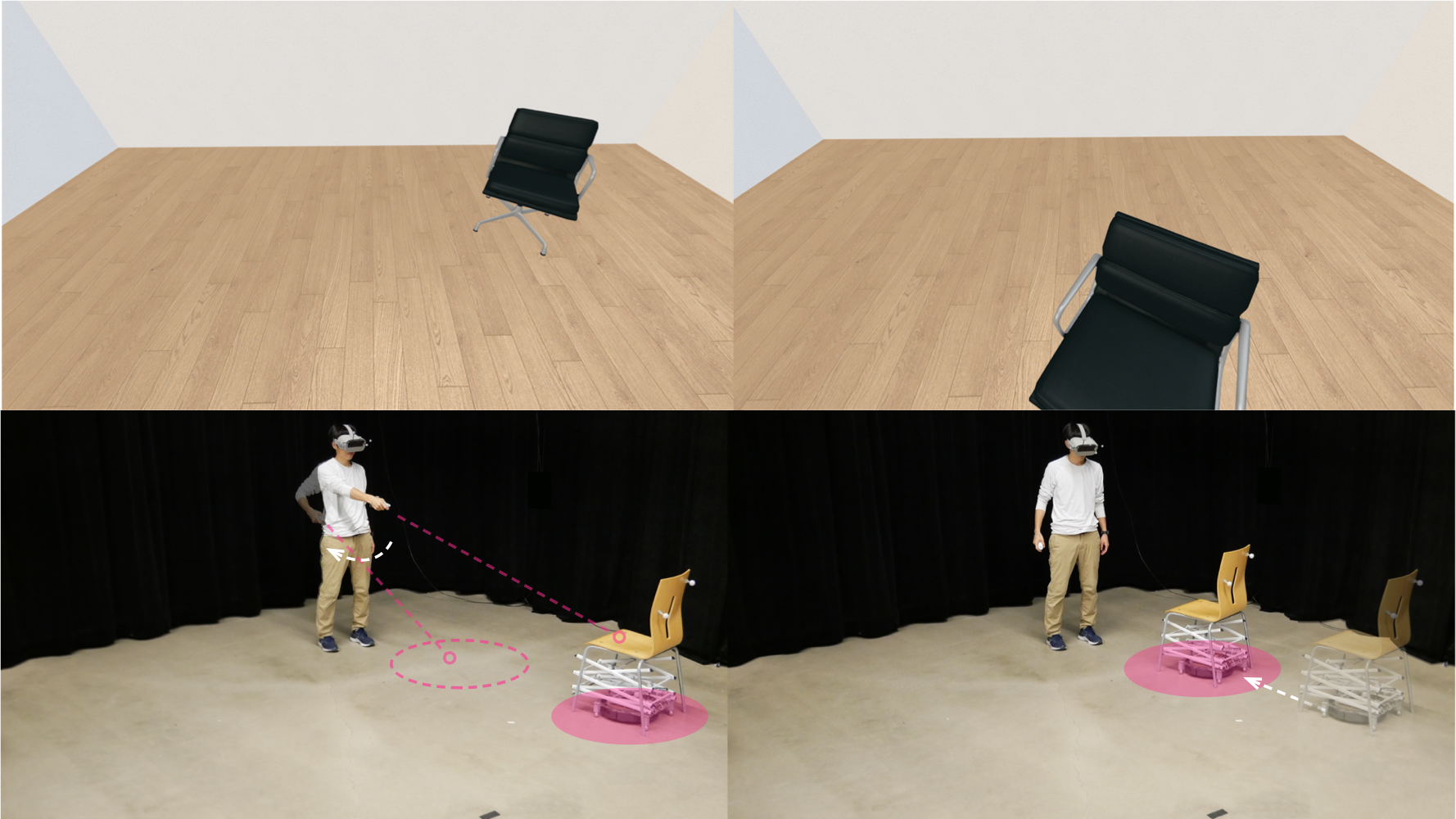}
\caption{Pointing and moving with a gesture.}
~\label{fig:interaction-grap}
\vspace{-0.4cm}
\end{figure}

\subsubsection{\changes{Virtual Scene Editing: Virtually Moving Furniture}}
\tool{} system also supports scene editing within VR. 
The virtual scene layout editing is similar to standard VR interactions and includes functionality like adding, removing, moving, resizing, rotating virtual building elements and furniture with a VR controller or a GUI.
For example, the user can point the controller at a virtual object and move it to a target location. The robot then updates the virtual object position (Figure~\ref{fig:interaction-grap}).



\changes{
\section{Discussion}
The reason for using 9 robots is based on our estimate of the maximum number of furniture and props that fit within close proximity (3-4 m range) of a single user in a standard VR scene - similar to how TurkDeck uses ten human actuators~\cite{cheng2015turkdeck}. By utilizing a small swarm of robots, we could also explore unique approaches to task coordination. A major limitation of past robotic graphics systems, where a single robot simulates multiple objects, is that the robot often moves too slowly to match fast graphics transformations. Swarm robots can coordinate and hand off tasks between each other to address this speed limitation. Two robots can, for instance, move two separate chairs into place to simulate a single chair rapidly moving across the room when teleporting. As the robots are not permanently attached to the scene objects, their role division can change depending on the virtual content. A single robot may at one point represent multiple objects, then switch roles to carry a copy of an object. Similar to other swarm robot systems, robots may also hand off tasks to recharge their batteries, providing a longer VR experience.

Utilizing props beyond furniture to represent a wider range of virtual scenes is also possible. For example, a combination of multiple pieces of furniture can represent different objects, such as a staircase simulated from stools of different heights. Moreover, the robots can move generic building blocks, custom cardboard props, or mannequins, which are useful for games such as Minecraft, or interaction design for human-robot communication. Beyond constructing static scenes, the system can also simulate dynamic objects and environments, such as height-changing desks, robotic beds, or transforming walls. This setup aids HCI researchers and practitioners in designing interactions for such emerging technologies.
}

%% file: 7-limitations.tex
\section{Limitations and Future Work}
There are many limitations of the current prototype system and opportunities for future work. 
First, due to the mechanical scissor lift design, the system can only actuate objects with a minimum 30 cm clearance underneath.
We tried different actuation mechanisms such as pushing or dragging the object, but this is only possible with lightweight objects (e.g., the robot was not able to drag a sofa). A more compact mechanical structure could alleviate this limitation. Second, the drivetrain of the robot is not omni-directional, therefore the orientation of the robot matters when picking up furniture, as it will influence in which direction the robot can move it without rotating it.

Currently, our mechanical scissor structure and wheeled-based robot are not robust enough to carry heavy objects and humans. However, by using a more robust scissor structure and stronger actuators, these shape-changing robots could also actuate furniture while the user is sitting or standing on. If this is possible, there are many more interesting application scenarios, such as simulating dynamic floor or terrain (e.g., CirculaFloor~\cite{iwata1999walking} or LevelUps~\cite{schmidt2015level}) or simulating dynamic objects.
In addition, such robots may be able to simulate the environment itself --- instead of bringing an existing chair, the robot changes its height to render a chair prop itself. This can introduce much more flexibility and dynamism for haptic sensations. We look forward to the future work which will investigate and demonstrate a more robust shape-changing structure. 

The feedback from our preliminary evaluation is encouraging, but a larger scale study would provide further insights into the appropriate utilization of swarm robots for VR haptics. 

In this paper, we focused on furniture rearrangement with wheel-based robots, but there are some intriguing alternative approaches. For example, recent research advances the capability of swarm construction for 3D architectures~\cite{jenett2017bill, werfel2014designing}.
Alternatively, we could also leverage different types of robots; for example, a swarm of drones~\cite{augugliaro2014flight} or cable-based robots~\cite{sato2002development} can provide mid-air haptic sensations. For future work, we are interested in exploring other types of swarm robots to enable more flexible spatial haptic interfaces.

Finally, while this paper entirely focused on haptics for a VR environment, \tool{} also has the potential for broader application space \changes{for dynamic office or home environments}. For example, these distributed robots can help the automation of home, labs, store, and public space by automatically reconfigure the spatial elements based on the situation (e.g., set up the meeting space, desk, and chair based on the calendar event, and clean up and reconfigure the space after the meeting.)
\tool{}'s capability of actuating existing objects is particularly interesting for this application space.
\changes{There are several technical challenges that would need to be addressed for such applications, including tracking methods (e.g., inside-out vs. outside-in tracking), interaction modalities (e.g., voice, gesture), and path planning (e.g., collision avoidance with multiple people).}
For example, the tracking system currently requires a dedicated setup, which is difficult to deploy outside a laboratory.
We are interested in reducing this required hardware setup, for example, using a single depth camera and AR markers to make the system more accessible.
Future work will investigate how to deploy these \tool{} robots into the real-world environment and investigate how these robots can be distributed and embedded in our physical environments and adapt to our everyday life.


%% file: 8-conclusion.tex
\section{Conclusion}
This paper introduces RoomShift, a room-scale dynamic haptic environment for virtual reality.
The goal of RoomShift is to provide room-scale haptic experiences. To achieve this goal, we propose the new approach of using a swarm of shape-changing robots that relocate existing furniture for reconfigurable physical environments. We described the design and implementation of RoomShift robot which leverages a wheel-based robot and expandable mechanical structure. The user evaluation study with five participants confirms the RoomShift's benefits in providing a realistic and enjoyable experience for the VR environment.
We demonstrate applications in virtual tours and architectural design and investigated several interaction techniques for these scenarios.

%% file: acknowledgements.tex
\section{Acknowledgements}
This research was partially supported by the Nakajima Foundation and the Strategic Information and Communications R\&D Promotion Programme (SCOPE) of the Ministry of Internal Affairs and Communications of Japan, and with equipment sponsored by the Engineering Excellence Fund at CU Boulder.